\pdfoutput=1

\documentclass[11pt]{article}

\usepackage[final]{acl}

\usepackage{times}
\usepackage{latexsym}

\usepackage[T1]{fontenc}

\usepackage[utf8]{inputenc}

\usepackage{microtype}

\usepackage{inconsolata}

\usepackage{graphicx}
\usepackage{enumitem}
\usepackage{amsmath}
\usepackage{multirow}
\usepackage{booktabs}
\usepackage{color}
\usepackage{amsmath}
\usepackage{xcolor}
\usepackage{colortbl}

\usepackage[normalem]{ulem}
\usepackage{array}
\newcolumntype{C}{>{\centering\arraybackslash}p{1.3cm}}
\newcolumntype{K}{>{\centering\arraybackslash}p{1.6cm}}

\definecolor{customblue}{RGB}{186,204,217}
\definecolor{yunfengwhite}{RGB}{216,227,231}

\usepackage{color, xspace}

\title{Bridging Relevance and Reasoning: Rationale Distillation in Retrieval-Augmented Generation}

\author{
    \textbf{Pengyue Jia\textsuperscript{1}\footnotemark[1]},
    \textbf{Derong Xu\textsuperscript{1,2}\footnotemark[1]},
    \textbf{Xiaopeng Li\textsuperscript{1}\footnotemark[1]},
    \textbf{Zhaocheng Du\textsuperscript{3}},
    \textbf{Xiangyang Li\textsuperscript{3}} \\
    \textbf{Yichao Wang\textsuperscript{3}},
    \textbf{Yuhao Wang\textsuperscript{1}},
    \textbf{Qidong Liu\textsuperscript{1}},
    \textbf{Maolin Wang\textsuperscript{1}},
    \textbf{Huifeng Guo\textsuperscript{3}}\\
    \textbf{Ruiming Tang\textsuperscript{3}\footnotemark[2]},
    \textbf{Xiangyu Zhao\textsuperscript{1}\footnotemark[2]}
\\
 \textsuperscript{1}City University of Hong Kong,\\
 \textsuperscript{2}University of Science and Technology of China,
 \textsuperscript{3}Huawei Noah's Ark Lab
\\
 \small{
   \textbf{Correspondence:} \href{mailto:email@domain}{xianzhao@cityu.edu.hk}
 }
}

\begin{document}
\maketitle
\renewcommand{\thefootnote}{\fnsymbol{footnote}}
\footnotetext[1]{Equal contribution}
\footnotetext[2]{Corresponding author}
\renewcommand{\thefootnote}{\arabic{footnote}}
\begin{abstract}
The reranker and generator are two critical components in the Retrieval-Augmented Generation (i.e., RAG) pipeline, responsible for ranking relevant documents and generating responses. However, due to differences in pretraining data and objectives, there is an inevitable misalignment between the documents ranked as relevant by the reranker and those required by the generator to support query-specific answers. To bridge this gap, we propose \textbf{RADIO}, a novel and practical preference alignment framework with \textbf{RA}tionale \textbf{DI}stillati\textbf{O}n. Specifically, we first propose a rationale extraction method that leverages the reasoning capabilities of Large Language Models (LLMs) to extract the rationales necessary for answering a query. Subsequently, a rationale-based alignment process is designed to rerank documents based on the extracted rationales and fine-tune the reranker to better align the preferences. Extensive experiments conducted on three tasks across four datasets demonstrate the effectiveness and transferability of our approach. Our code is released online\footnote{\url{https://github.com/Applied-Machine-Learning-Lab/RADIO}}.
\end{abstract}

\section{Introduction}

Large Language Models (LLMs), pretrained on massive datasets, have demonstrated exceptional reasoning and text generation capabilities, as evidenced by prior research~\cite{zhao2023survey,ni2025zeroed,wang2023large,xu2024multi,fu2023unified}. These models also adhere to the scaling laws, exhibiting improvements in performance and intelligence as the number of parameters increases~\cite{kaplan2020scaling}. Retrieval-augmented generation (RAG)~\cite{xu2025align,xu2025harnessing} builds upon these capabilities by integrating information retrieval mechanisms~\cite{zhao2018deep,zhao2018recommendations,liu2024large} with generative models, such as LLMs. This approach not only mitigates the problem of hallucination in text generation but also enhances the system's adaptability to dynamically evolving information needs, making it a robust solution for tasks requiring both accuracy and contextual relevance~\cite{gao2023retrieval}.

However, RAG pipelines typically assemble components (e.g., the reranker and generator~\cite{fan2024survey}) that have been pretrained separately. Due to differences in their pretraining data and optimization objectives, these components often exhibit varying preferences, which can impact the overall effectiveness of the system. Specifically, pretrained rerankers~\cite{bge_embedding} are designed to evaluate the relevance between queries and documents. However, the documents identified as relevant under this criterion may not provide the necessary support for reasoning to derive an accurate answer to the query. Bridging this gap between the reranker’s relevance measurement and the generator’s reasoning requirements presents a significant challenge that must be addressed to improve the RAG pipeline's performance.

Recent studies try to address this gap by training a bridge model~\cite{ke2024bridging}, using LLM-based scores~\cite{zhang2024arl2} or combining both LLM-based and retrieval-based scores~\cite{dong2024understand} to fine-tune RAG components. Additionally, while some other methods are not explicitly designed for this problem, they can indirectly contribute to bridging the gap. These approaches can use response quality~\cite{ma2023query} or perplexity distillation~\cite{izacard2023atlas,shi2023replug} as signals to fine-tune the reranker.
Despite showing promise, these approaches face critical limitations: their alignment signals rely solely on the surface-level connection between the query/answer and document, failing to capture the deeper reasoning processes or more complex relationships involved.

To address the above limitation, we propose RADIO, a novel and practical preference alignment framework with rationale distillation in RAG. RADIO leverages rationale as a signal to bridge the reranker's relevance measurement with the generator's reasoning requirements for response generation.
First, to efficiently extract the rationales needed to answer a query, we use the query and its ground truth answer as context and generate the rationales with LLMs.
Second, to mitigate the preference misalignment between the reranker and generator while ensuring the solution remains practical, we rerank the documents based on the extracted rationales and fine-tune the reranker. This step distillates rationales from generators to rerankers, and aligns the reranker with the generator’s information needs for answering the query effectively.

\begin{figure*}
    \centering
    \includegraphics[width=\textwidth]{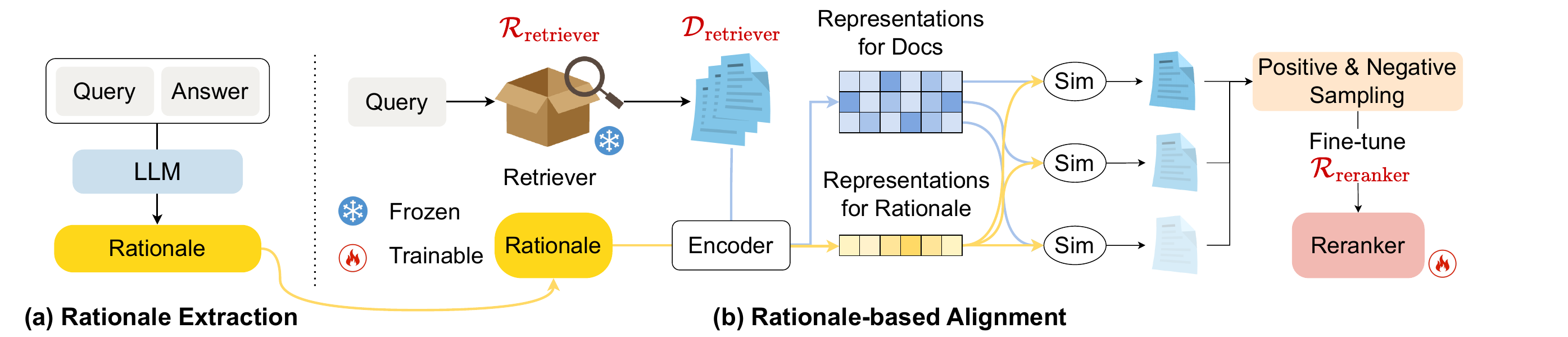}
    \caption{Overview of RADIO.}
    \label{fig:overview}
\end{figure*}

RADIO effectively addresses the preference inconsistency between RAG components by first generating a comprehensive rationale and then fine-tuning the reranker based on the extracted rationale. This approach considers the deeper reasoning behind answers. We evaluate RADIO on three tasks across four datasets: Open-domain QA (NQ~\cite{kwiatkowski2019natural} and TriviaQA~\cite{joshi2017triviaqa}), Multi-choice questions (MMLU~\cite{hendrycks2020measuring}), and Multi-hop QA (Musique~\cite{trivedi2022musique}). The results validate the superiority of our method compared to other state-of-the-art baselines. Our contributions can be summarized as follows:

\begin{itemize}[leftmargin=*]
    \item We propose RADIO, a novel and practical framework designed to address the preference misalignment of different components in RAG pipelines. 
    \item We introduce rationale distillation within the RAG framework, which is an effective approach that leverages explicit textual rationales as signals to align the preferences of different components in RAG.  
    \item Extensive experiments are conducted on three tasks across four datasets to demonstrate the effectiveness and transferability of RADIO.  
\end{itemize}

\section{Related Work}

\textbf{Retrieval-Augmented Generation.}
Large language models (LLMs) have demonstrated groundbreaking performance across numerous tasks~\cite{jia2025agentir} but still face challenges such as hallucination and outdated knowledge~\cite{gao2023retrieval}. To address these issues, retrieval-augmented generation (RAG) has been introduced~\cite{fan2024survey,zhang2025lsrp,jia2024g3}. RAG retrieves relevant information from external knowledge bases and incorporates it as contextual input to the generator (LLM), enhancing the accuracy and reliability of the generated responses. The typical RAG pipeline can be divided into several key components: query rewriter, retriever, reranker, and generator. The query rewriter~\cite{wang2023query2doc} modifies and expands~\cite{jia2024mill,li2023agent4ranking} the original query to improve retrieval recall, ensuring more relevant documents are retrieved. The retriever~\cite{bge-m3} fetches relevant documents based on the query. Dense retrievers generally outperform sparse retrievers in this step. To integrate contextual information more effectively and identify documents more relevant to the query, rerankers~\cite{moreira2024enhancing} with larger models and greater complexity are introduced to reorder the retrieved documents compared to retrievers. Finally, the generator—usually a powerful LLM-uses the query and the top-$k$ documents from the reranker to generate the final response. 
In this work, we address the issue of preference misalignment among different components within the RAG pipeline. We aim to leverage rationale as a signal to align these preferences and enhance the overall performance of the RAG system. 
It is worth noting that RADIO and methods like ReAct~\cite{yao2022react} are not mutually exclusive and can be used together to complement each other. 
While RADIO enhances the ranking of retrieved documents at each step, ReAct focuses on task-solving through reasoning and action.

\noindent\textbf{Preference Alignment.}
To further improve LLMs, preference alignment is often performed after the initial pretraining phase~\cite{jiang2024survey}. Approaches such RLHF~\cite{ouyang2022training} and DPO~\cite{rafailov2024direct} are proposed to align the output of LLMs more closely with human preferences.
DPO transforms tasks into classification problems, achieving high computational efficiency and strong performance. 
In the context of RAG, several works~\cite{ke2024bridging} can be transformed to address the challenge of preference alignment between RAG components. REPLUG~\cite{shi2023replug} improves RAG pipelines involving black-box LLMs by using the probability of the LLM generating the correct answer as a signal to determine document importance. Similarly, RRR~\cite{ma2023query} uses metrics based on the quality of the LLM's generated response as a signal to evaluate a document’s utility. On the other hand, ARL2~\cite{zhang2024arl2} prompts LLMs to generate self-guided relevance labels for fine-tuning retriever, and DPA-RAG~\cite{dong2024understand} introduces a bidirectional alignment strategy to mitigate preference inconsistencies in RAG pipelines. In this work, we focus on optimizing the reranker within RAG. Our goal is to enable the reranker to effectively identify supportive documents well-suited for the generator, whether black-box or open-source, facilitating the production of accurate outputs. 
In addition, BGM~\cite{ke2024bridging} trains a bridge model between retriever and LLMs to transform the retrieved information into the format LLM's prefer. Our method is theoretically compatible with BGM and could be combined to jointly enhance performance without any conflict.

\section{Methodology}

In this section, we detail rationale distillation in RAG. Specifically, we first demonstrate the task definition of RAG in Section~\ref{sec:task_definition}. Then we give an overview of our proposed framework in Section~\ref{sec:overview}, introduce the rationale extraction method in Section~\ref{sec:rationale_extraction}, and detail the rationale-based alignment in Section~\ref{sec:alignment} and optimization in  Section~\ref{sec:opt}.

\subsection{Task Definition} \label{sec:task_definition}

To address the hallucination problem and enhance adaptability to dynamic information of LLMs, RAG systems have been proposed. These systems enhance generative models by introducing additional contextual information retrieved based on a given query $q$. Specifically, when a query $q$ is input into the RAG pipeline, the retriever $\mathcal{R}_{\text{retriever}}$ first retrieves relevant documents by calculating similarity scores and top-$k_1$ selection. The process can be formalized as follows:
\begin{equation}
    \mathcal{D}_\text{retriever}= \{\boldsymbol{d}_i\mid \boldsymbol{d}_i\in\text{Top-}k_1(\text{score}_{\text{retriever}}(q,d))\}
\end{equation}
where $q$ and $d$ are the query and document, $\text{score}_{\text{retriever}}$ denotes the score function in retriever, $\boldsymbol{d}_i$ means the $i$-th document in corpus and $\mathcal{D}_\text{retriever}$ is the documents set output by retriever $\mathcal{R}_{\text{retriever}}$.

To eliminate contextually irrelevant noise and provide more precise contextual information, the initially filtered documents will be further reranked by the reranker:
\begin{align}
    \mathcal{D}_\text{reranker}= \{\boldsymbol{d}_j\mid \boldsymbol{d}_j\in\text{Top-}k_2(\text{score}_{\text{reranker}}(q,d))\}
\end{align}
where $\mathcal{D}_\text{reranker}$ is the documents selected by reranker, $\text{score}_{\text{reranker}}$ denotes the score function in reranker. Note that the $j$th document $\boldsymbol{d}_j$ is in $\mathcal{D}_{\text{retriever}}$ (i.e., $d_j \in \mathcal{D}_{\text{retriever}}$) and $k_2$ is the number of documents selected by reranker, which is smaller than $k_1$ used by the retriever.

Finally, the documents filtered by the reranker, along with the original query, will be fed into the generator as contextual information to help generate the final response:
\begin{equation}
    \hat{y} = \mathcal{G}(q, \mathcal{D}_\text{reranker})
\end{equation}
where $\hat{y}$ is the generated response and $\mathcal{G}$ denotes the generator.

It is worth noting that the documents selected by the reranker directly influence the generator's input. Therefore, in this work, we aim to align the preferences of the reranker and generator to enhance their consistency. This alignment improves the overall accuracy of the RAG system's responses.

\subsection{Framework Overview} \label{sec:overview}

The overview of RADIO is depicted in Figure~\ref{fig:overview}. RADIO is consisted of two phases: rationale extraction (Figure~\ref{fig:overview}(a)) and rationale-based alignment (Figure~\ref{fig:overview}(b)).

In the rationale extraction process, we combine the query with its ground truth answer and input them into LLMs to generate precise rationales. Using the correct answer as context in the prompt improves the accuracy of the LLM’s rationale generation, ensuring that the generated rationale closely aligns with the requirements for deriving the correct answer.

In the rationale-based alignment process, our goal is to use the generated rationale to guide rerankers, enabling it to select documents that better support the generator in answering the query. Specifically, we leverage the generated rationale as a signal to rerank documents. The reranked documents will be used to fine-tune the reranker, addressing the preference misalignment between the reranker and the generator. By aligning these components, the process ensures that the selected documents are not only contextually relevant but also optimally supportive for the generator's reasoning and response generation.

\subsection{Rationale Extraction} \label{sec:rationale_extraction}

Rationales are critical components of LLM reasoning processes and have been shown to enhance the accuracy of LLM-generated responses significantly. This perspective is supported by existing works such as Chain-of-Thoughts (CoT~\cite{wei2022chain}). Existing work~\cite{shi2023replug,ma2023query,dong2024understand} has primarily focused on the initial relationships between queries and documents or indirect relationships between answers and documents, while overlooking rationales, a crucial intermediary component in the reasoning process. Motivated by this, we aim to extract rationales as signals to align the preferences of different components in the RAG pipeline. 

To accurately extract the rationale necessary for answering a query and deriving the correct answer, we combine the query with the ground truth answer as contextual information of LLMs, as shown in Figure~\ref{fig:overview}(a). The prompt template used for this process is as follows: "\textit{You are a professional QA assistant. Given a question and the ground truth answer, you can output the rationale why the ground truth answer is correct. Question: \{question\}. Answer: \{answer\}. Rationale: }". The generation process can be formalized as:
\begin{equation}
    r = \text{LLM}(q,a)
\end{equation}
where $q$ and $a$ are the query and answer, $r$ deontes the generated rationale.

By doing so, we effectively bridge the gap between the query and the answer by generating the necessary rationale. This rationale accurately supports the reasoning process required to derive the correct answer from the query.

\subsection{Rationale-based Alignment} \label{sec:alignment}

Given the extracted rationale $r$, a key challenge lies in effectively and efficiently utilizing it to improve preference consistency within the RAG pipeline. In this section, we propose a rationale-based alignment approach, where the rationale serves as a signal to fine-tune the reranker. This enables the reranker to identify and prioritize supportive documents that facilitate the generator in producing accurate responses. Specifically, we first use the retriever $\mathcal{R}_{\text{retriever}}$ to retrieve $k_1$ relevant documents based on the query: 
\begin{equation}
    \mathcal{D}_{\text{retriever}}=\mathcal{R}_{\text{retriever}}(q, \mathbf{\mathcal{C})}
\end{equation}
where $\mathcal{D}_{retriever}$ is the document set retrieved by $\mathcal{R}_{\text{retriever}}$ based on query $q$ and corpus $\mathbf{\mathcal{C}}$, and $|\mathcal{D}_{retriever}|=k_1$.

Next, to facilitate the comparison of similarity between different documents and the rationale, we use a text encoder to convert both the documents and the rationale into dense vectors.
\begin{align}
    \boldsymbol{e}^{\text{document}}_i = \text{Encoder}(\boldsymbol{d}_i) \\
    \boldsymbol{e}^{\text{rationale}} = \text{Encoder}(r)
\end{align}
where $\boldsymbol{e}^{\text{document}}_i$ and $\boldsymbol{e}^{\text{rationale}}$ denote the representations of $i$th document and rationale. $\boldsymbol{d}_i$ is the $i$th document and $r$ represents the extracted rationale.

Then, we calculate the semantic similarity between each document and the rationale. Here, we use cosine similarity, denoted as $\text{sim}(\cdot)$. The calculated scores indicate the degree to which each document supports generating the correct answer, with higher scores reflecting stronger support. We also linearly interpolate the score of documents with their retrieval score in the retrieval stage by weighted score sum.
\begin{gather}
    \boldsymbol{s}_i^{\text{rationale}} = \text{sim}(\boldsymbol{e}^{\text{document}}_i, \boldsymbol{e}^{\text{rationale}}) \\
    \boldsymbol{s}_i^{\text{retriever}} = \text{score}_{\text{retriever}}(q, \boldsymbol{d}_i) \\
    \boldsymbol{s}_i^{\prime} = \alpha \boldsymbol{s}_i^{\text{rationale}} + (1-\alpha)\boldsymbol{s}_i^{\text{retriever}}
\end{gather}
where $\boldsymbol{s}_i^{\text{rationale}}$, $\boldsymbol{s}_i^{\text{retriever}}$, $\boldsymbol{s}_i^{\prime}$ represent the rationale similarity score, retrieval score, and final score for $i$th document. $\text{score}_{\text{retriever}}(\cdot)$ is the score function in retriever and $\alpha$ is a hyperparameter used for integration. Note that we apply min-max normalization in this work to both rationale score and retrieval score before integration.

Next, we rerank the documents based on their scores. Following the previous sampling method \textit{Top-k shifted by N}~\cite{moreira2024nv}, we select the top-ranked document as the positive sample and then shift by $n$ documents and sample $m$ negative samples from the subsequent documents to construct positive-negative pairs for fine-tuning the reranker. This process can be represented as:
\begin{gather}
    \boldsymbol{d}_{\text{pos}} = \boldsymbol{d}_i, \text{where} \  i=\arg\max_{i}\boldsymbol{s}_i^{\prime} \\
    \{\boldsymbol{d}_{\text{neg}}\}=\text{Sample}_m(\{\boldsymbol{d}_i|\text{rank}(\boldsymbol{d}_i) \textgreater n \})
\end{gather}
where $\boldsymbol{d}_{\text{pos}}$ and $\boldsymbol{d}_{\text{neg}}$ are the sampling positive and negative documents, $\text{Sample}_m(\cdot)$ denotes a sampling operation that selects $m$ negative documents from the set of documents ranked lower than $n$.

\subsection{Optimization} \label{sec:opt}
Following BGE embedding~\cite{bge_embedding} and QA Ranking Benchmark~\cite{moreira2024enhancing}, we use InfoNCE as our optimization objectives to fine-tune reranker:
\begin{gather}
    f(q,d)=\text{exp}(\phi(q,d)/\tau) \\
    L = -\text{log}\frac{f(q, d^+)}{f(q,d^+) + \sum_{i=1}^{N} f(q,d_i^-)}
\end{gather}
where $d^+$ and $d^-$ represent the positive and negative document, $\tau$ is the temperature parameter, and $N$ denotes the number of negative documents.

\section{Experiments}

\subsection{Datasets and Metrics}

Following previous work~\cite{shi2023replug,zhang2024arl2}, we conduct experiments on three tasks across four datasets to evaluate RADIO with other methods: \textbf{Open-domain QA} (NQ~\cite{kwiatkowski2019natural} and TriviaQA~\cite{joshi2017triviaqa}), \textbf{Multi-choice questions} (MMLU~\cite{hendrycks2020measuring}), and \textbf{Multi-hop QA} (Musique~\cite{trivedi2022musique}). Detailed dataset descriptions are given in Appendix~\ref{sec:appendix_dataset}.
Following previous work~\cite{ma2023query,shi2023replug}, we report EM and F1 scores for QA datasets and EM for MMLU.

\subsection{Baselines}

To verify the effectiveness of RADIO, we conduct experiments with the following baseline methods: \textbf{Base}~\cite{bge_embedding}, \textbf{Atlas}~\cite{izacard2023atlas},  \textbf{REPLUG}~\cite{shi2023replug}, \textbf{Trainable rewrite-retrieve-read (RRR)}~\cite{ma2023query}, \textbf{ARL2}~\cite{zhang2024arl2}, and \textbf{DPA-RAG}~\cite{dong2024understand}. The detailed introduction of baselines is given in Appendix~\ref{sec:appendix_baseline}.

\begin{table*}[t]
\centering
\caption{Overall experiments. ``\textbf{{\large *}}'' indicates the statistically significant improvements (i.e., two-sided t-test with $p<0.05$) over the best baseline. For all metrics, higher is better. $\Delta$ represents the relative improvement of RADIO over Base method.}
\label{table:overall}
\resizebox{\textwidth}{!}{
\begin{tabular}{c|cccccc|cccccc} 
\toprule
\multirow{3}{*}{Method} & \multicolumn{6}{c|}{NQ}                                                                                        & \multicolumn{6}{c}{TriviaQA}                                                                                   \\ 
\cmidrule{2-13}
                        & \multicolumn{2}{c}{gte-base}      & \multicolumn{2}{c}{gte-large}     & \multicolumn{2}{c|}{bge-reranker-base} & \multicolumn{2}{c}{gte-base}      & \multicolumn{2}{c}{gte-large}     & \multicolumn{2}{c}{bge-reranker-base}  \\ 
\cmidrule{2-13}
                        & EM              & F1              & EM              & F1              & EM              & F1                   & EM              & F1              & EM              & F1              & EM              & F1                   \\ 
\midrule
Base                    & 0.2931          & 0.4046          & 0.2798          & 0.3935          & 0.3371          & 0.4603               & 0.5449          & 0.6374          & 0.5495          & 0.6414          & 0.6114          & 0.7120               \\
Atlas                   & 0.3338          & 0.4587          & 0.3418          & 0.4677          & 0.3521          & 0.4832               & 0.5823          & 0.6752          & 0.6004          & 0.6972          & 0.6083          & 0.7063               \\
REPLUG                  & 0.3257          & 0.4484          & 0.2607          & 0.3670          & 0.3427          & 0.4753               & 0.5679          & 0.6578          & 0.5248          & 0.6171          & 0.6032          & 0.7004               \\
RRR                     & 0.3374          & 0.4608          & 0.3299          & 0.4578          & 0.3438          & 0.4754               & 0.5801          & 0.6716          & 0.5358          & 0.6237          & 0.6099          & 0.7091               \\
ARL2                    & \uline{0.3413}  & \uline{0.4688}  & \uline{0.3515}  & \uline{0.4804}  & \uline{0.3568}  & \uline{0.4885}       & 0.6079          & \uline{0.7086}  & \uline{0.6107}  & \uline{0.7120}  & 0.6137          & 0.7149               \\
DPA-RAG                 & 0.3391          & 0.4674          & 0.3385          & 0.4710          & 0.3462          & 0.4793               & \uline{0.6080}  & 0.7076          & 0.6097          & 0.7119          & \uline{0.6149}  & \textbf{0.7169*}      \\
RADIO (Ours)            & \textbf{0.3512*} & \textbf{0.4790*} & \textbf{0.3565*} & \textbf{0.4850*} & \textbf{0.3665*} & \textbf{0.4917*}      & \textbf{0.6084} & \textbf{0.7095*} & \textbf{0.6128*} & \textbf{0.7137*} & \textbf{0.6154} & \uline{0.7151}       \\
\rowcolor{yunfengwhite} $\Delta$ & $\uparrow19.82$\%& $\uparrow18.39$\%& $\uparrow$27.41\%& $\uparrow$23.25\%& $\uparrow$8.72\%& $\uparrow$6.82\%& $\uparrow$11.55\%& $\uparrow$11.31\%& $\uparrow$11.52\%& $\uparrow$11.27\%& $\uparrow$0.65\%& $\uparrow$0.40\%\\ 
\bottomrule
\end{tabular}}
\end{table*}

\begin{table}
\centering
\caption{Experimental results on MMLU. EM is reported as the evaluation metric. The source dataset used to fine-tune rerankers is the Open-domain QA dataset NQ. $\Delta$ represents the relative improvement of RADIO over Base method.}
\label{table:mmlu}
\resizebox{\linewidth}{!}{
\begin{tabular}{c|CCCCC} 
\toprule
Method       & Humanities      & Social          & STEM            & Other           & ALL              \\ 
\midrule
Base         & 0.4089          & 0.6867          & \textbf{0.5147} & 0.6650          & 0.5502           \\
Atlas        & 0.3985          & 0.6935          & 0.5074          & 0.6563          & 0.5447           \\
REPLUG       & 0.4102          & 0.6854          & 0.5065          & 0.6590          & 0.5473           \\
RRR          & 0.4079          & 0.6913          & \uline{0.5116}  & 0.6572          & 0.5484           \\
ARL2         & 0.4147          & \uline{0.7016}  & 0.5106          & 0.6630          & 0.5540           \\
DPA-RAG      & \uline{0.4157}  & 0.701           & 0.5078          & \uline{0.6652}  & \uline{0.5541}   \\
RADIO (Ours) & \textbf{0.4172} & \textbf{0.7013} & 0.5080          & \textbf{0.6717} & \textbf{0.5562}  \\
\rowcolor{yunfengwhite} $\Delta$ & $\uparrow$2.03\% & $\uparrow$2.13\% & $\downarrow$1.30\% & $\uparrow$1.01\%  & $\uparrow$1.09\%\\
\bottomrule
\end{tabular}}
\end{table}

\subsection{Backbone Rerankers}

To validate the generality and adaptability of RADIO, we select three different rerankers as the backbone models for our experiments: \textbf{gte-base}~\cite{li2023towards}, \textbf{gte-large}~\cite{li2023towards}, and \textbf{bge-reranker-base}~\cite{bge_embedding}. A detailed introduction is given in Appendix~\ref{sec:appendix_reranker}.

\subsection{Implementation Details}

We implement RADIO on FlashRAG~\cite{jin2024flashrag}, a Python library for efficient RAG research. In the RAG pipeline, we take \textit{e5-base-v2}~\cite{wang2022text} as the retriever, and \textit{Meta-Llama-3.1-8B-Instruct}~\cite{touvron2023llama} as the generator. We sample 20,000 instances from NQ and TriviaQA to construct the fine-tuning dataset and fine-tune rerankers separately. For document sampling, we set the shift $n$ in \textit{Top-k shifted by N} method as 3, and sample 6 negative samples from the subsequent documents. To ensure a fair comparison, the sampling index is fixed and remains unchanged across methods. In the RAG pipeline, we set the number of documents selected by retriever and reranker (i.e., $k_1$ and $k_2$) as 20 and 5. The retrieval corpus we used in experiments is Wikipedia (2018, December). For fine-tuning the reranker, we tune the training epochs from 1 to 5 and the integration hyperparameter $\alpha$ from 0.0 to 1.0. We use Adam~\cite{kingma2014adam} optimizer with a learning rate 6e-5 and a weight decay of 0.01. The prompts we used in experiments are given in Appendix~\ref{sec:appendix_prompts}. We provide an analysis of the impact of fine-tuning sample size on performance in Appendix~\ref{sec:appendix_sample}, and we also give a detailed analysis on the shift parameter $n$ and the negative samples $m$ in Appendix~\ref{sec:appendix_m_n}.

\subsection{Main Results}

\begin{table*}
\centering
\caption{Transferability analysis across generators. EM is reported as the metric for MMLU dataset.}
\label{table:trans}
\resizebox{0.95\textwidth}{!}{
\begin{tabular}{c|c|cc|KKKKK} 
\toprule
\multirow{2}{*}{Generator}            & \multirow{2}{*}{Method} & \multicolumn{2}{c|}{NQ} & \multicolumn{5}{c}{MMLU}                        \\ 
\cmidrule{3-9}
                                      &                         & EM     & F1             & Humanities & Social & STEM   & Other  & ALL     \\ 
\midrule
\multirow{2}{*}{Llama3.1-8b-instruct} & Base                    & 0.3371 & 0.4603         & 0.4089     & 0.6867 & \textbf{0.5147} & 0.6650 & 0.5502  \\
                                      & RADIO (ours)          & \textbf{0.3665} & \textbf{0.4917}         & \textbf{0.4172}     & \textbf{0.7013} & 0.5080 & \textbf{0.6717} & \textbf{0.5562}  \\ 
\midrule
\multirow{2}{*}{qwen2.5-14b-instruct} & Base                    & 0.3310 & 0.4484         & 0.5439     & 0.8200 & 0.7206 & 0.7541 & 0.6906  \\
                                      & RADIO (ours)          & \textbf{0.3518} & \textbf{0.4753}         & \textbf{0.5598}     & \textbf{0.8229} & \textbf{0.7250} & \textbf{0.763}1 & \textbf{0.6995}  \\ 
\midrule
\multirow{2}{*}{CLM}           & Base                    & 0.3607 & 0.4880         & 0.6485     & 0.8362 & \textbf{0.6784} & 0.8005 & 0.7300  \\
                                      & RADIO (ours)          & \textbf{0.3742} & \textbf{0.5086}         & \textbf{0.6548}     & \textbf{0.8372} & 0.6768 & \textbf{0.8010} & \textbf{0.7321}  \\
\bottomrule
\end{tabular}}
\end{table*}

\subsubsection{Open-domain QA}

To evaluate the effectiveness of RADIO and its transferability across different rerankers, we conduct experiments on the NQ and TriviaQA datasets. The results are presented in Table~\ref{table:overall}. A case study is given in Appendix~\ref{sec:appendix_case}. From these results, we can draw the following conclusions:
\begin{itemize}[leftmargin=*]
    \item Compared to the Base method, most experimental settings achieve better results, demonstrating the necessity of preference alignment within the RAG pipeline. 
    \item Compared to other baseline methods, RADIO consistently achieves superior performance across all datasets and reranker backbone configurations. The results validate the effectiveness of using rationales as signals for preference alignment in RAG pipeline.  
    \item On TriviaQA, methods such as RRR and REPLUG show performance declines relative to the base method when using rerankers \textit{gte-large} and \textit{bge-reranker-base}. This indicates that these methods are sensitive, limiting their applicability. In contrast, RADIO demonstrates robust adaptability to different rerankers, achieving significant performance improvements across all three rerankers.  
    \item As the reranker becomes larger or more powerful (e.g., progressing from \textit{gte-base} to \textit{gte-large} and further to \textit{bge-reranker-base}), the performance ranking of models fine-tuned with RADIO aligns with the reranker’s inherent capabilities. This suggests that RADIO's performance gains are sustainable and scalable with stronger rerankers, providing an avenue to further explore the upper performance limits of RAG pipelines.
\end{itemize}

\subsection{Task Generalization Analysis}

\subsubsection{Multi-choice Questions}
We also conduct experiments on MMLU. Since MMLU is a multiple-choice dataset, we report the EM metric~\cite{ma2023query}. Additionally, following previous work~\cite{yu2023augmentation}, we fine-tune the reranker \textit{bge-reranker-base} using open-domain QA as the source task and evaluate its performance on the MMLU dataset. Table~\ref{table:mmlu} shows the results of fine-tuning reranker with NQ dataset. The results of fine-tuning reranker with TriviaQA are given in Appendix~\ref{sec:appendix_mmlu}. We can draw the following conclusions:
\begin{itemize}[leftmargin=*]
    \item From the metrics corresponding to the ALL category, RADIO demonstrates consistent improvements over Base. This highlights the effectiveness and generalization of RADIO, as it successfully adapts to multi-choice question tasks even when fine-tuned on the Open-domain QA tasks.
    \item Analyzing the results by question category, RADIO shows more significant improvements over the Base method in the Humanities and Social Sciences categories, with average gains of \underline{\textbf{2.03\%}} and \underline{\textbf{2.13\%}}, respectively. However, it exhibits a slight negative effect in the STEM category. This may be due to the fine-tuning datasets (NQ and TriviaQA), which are Open-domain QA datasets with distributions more similar to humanities and social sciences but markedly different from STEM subjects.
    \item Compared to other baseline methods, RADIO achieves top performance in the vast majority of metrics, demonstrating its superiority and state-of-the-art capability.
\end{itemize}

\subsubsection{Multi-hop QA}

\begin{table}
\centering
\caption{Experimental results on Musique. EM and F1 are reported as the metrics. The source datasets used to fine-tune rerankers are Open-domain QA datasets NQ and TriviaQA. $\Delta$ represents the relative improvement of RADIO over Base method.}
\label{table:musique}
\resizebox{\linewidth}{!}{
\begin{tabular}{c|cccc} 
\toprule
\multirow{2}{*}{Method} & \multicolumn{2}{c}{NQ (Source Dataset)} & \multicolumn{2}{c}{TriviaQA (Source Dataset)}  \\ 
\cmidrule{2-5}
                        & EM              & F1                    & EM              & F1                           \\ 
\midrule
Base                    & \uline{0.0662}  & 0.1297                & 0.0662          & 0.1297                       \\
Altlas                  & 0.0658          & \uline{0.1336}        & 0.0541          & 0.1184                       \\
REPLUG                  & 0.0608          & 0.1269                & 0.0559          & 0.1188                       \\
RRR                     & 0.0658          & 0.1315                & 0.0592          & 0.1207                       \\
ARL2                    & 0.0641          & 0.1309                & 0.0674          & 0.1330                       \\
DPA-RAG                 & 0.0629          & 0.1307                & \uline{0.0690}  & \uline{0.1337}               \\
RADIO (Ours)            & \textbf{0.0699} & \textbf{0.1371}       & \textbf{0.0695} & \textbf{0.1347}              \\
                 \rowcolor{yunfengwhite}    $\Delta$   & $\uparrow$5.59\%            & $\uparrow$5.71\%                  & $\uparrow$4.98\%            & $\uparrow$3.86\%                         \\
\bottomrule
\end{tabular}}
\end{table}

We also conduct experiments on the multi-hop dataset Musique with reranker \textit{bge-reranker-base}, and the experimental results are shown in Table~\ref{table:musique}. Based on the experimental results, we can draw the following conclusions:
\begin{itemize}[leftmargin=*]
    \item First, RADIO achieves significant performance improvements in all evaluation metrics (EM and F1), demonstrating its task generalization ability in multi-hop reasoning tasks. On the NQ source dataset, RADIO achieves an EM value of 0.0699 and an F1 value of 0.1371, with improvements of 5.59\% and 5.71\% over the Base method, respectively. On the TriviaQA source dataset, RADIO's EM value is 0.0695, and F1 value is 0.1347, showing improvements of 4.98\% and 3.86\% over the Base method.
    \item Secondly, compared to other baseline methods, RADIO demonstrates a more balanced and consistent advantage across both the NQ and TriviaQA source datasets. Notably, when using NQ as the source dataset, only RADIO shows a positive improvement in the EM metric, while all other baseline methods experience varying degrees of negative degradation.
\end{itemize}

\subsection{Transferability Analysis across Generators}

We conduct experiments on two datasets of different tasks, NQ and MMLU, using three different generators (\textit{Llama3.1-8b-instruct}~\cite{touvron2023llama}, \textit{qwen2.5-14b-instruct}~\cite{qwen2}, and an advanced closed-source LLM (CLM) with \textit{bge-reranker-base} fine-tuned on NQ to validate the transferability of our method, as shown in Table~\ref{table:trans}. We can find that (1) RADIO maintains its effectiveness across different generators, consistently enhancing the performance of the original RAG pipeline. This demonstrates RADIO's strong transferability with various generators. (2) Comparing different generators reveals that RADIO's performance gains are more pronounced with smaller and less capable generators. Specifically, when the generators are Llama3.1, Qwen2.5, and CLM, RADIO achieves EM improvements of 8.72\%, 6.28\%, and 3.74\%, respectively, and F1 improvements of 6.82\%, 6.00\%, and 4.22\%. This is because as the generator's capability increases and approaches the upper performance limits of the RAG pipeline, further enhancing the pipeline becomes increasingly challenging, resulting in a smaller improvement.

\subsection{Ablation Study}

\begin{table}
\centering
\caption{Ablation study.}
\label{table:ablation}
\resizebox{\linewidth}{!}{
\begin{tabular}{ccccc} 
\toprule
Dataset               & Metrics         & w/o ALL         & w/o Retrieval  & RADIO          \\ 
\midrule
\multirow{2}{*}{NQ}   & EM              & 0.3371          & \uline{0.3587} & \textbf{0.3665}  \\
                      & F1              & 0.4603          & \uline{0.4858} & \textbf{0.4917}  \\ 
\midrule
\multirow{5}{*}{MMLU} & Humanities (EM) & 0.4089          & \uline{0.4168} & \textbf{0.4172}  \\
                      & Social (EM)     & 0.6867          & \uline{0.6981} & \textbf{0.7013}  \\
                      & STEM (EM)       & \textbf{0.5147} & \uline{0.5109} & 0.508            \\
                      & Other (EM)      & 0.665           & \uline{0.6666} & \textbf{0.6717}  \\
                      & ALL (EM)        & 0.5502          & \uline{0.5548} & \textbf{0.5562}  \\
\bottomrule
\end{tabular}}
\end{table}

To explore the specific impact of rationale and retrieval score, we design the following variants with reranker \textit{bge-reranker-base}: 
\begin{itemize}[leftmargin=*]
    \item \textbf{w/o ALL:} Base reranker without fine-tuning. Do not introduce rationale or retrieval score.
    \item \textbf{w/o Retrieval:} Ranking documents and fine-tuning reranker only based on the rationale scores.
    \item \textbf{RADIO:} Fine-tuning reranker based on both rationale and retrieval scores.
\end{itemize}

Table~\ref{table:ablation} shows the results of ablation study on NQ and MMLU, where we can derive the following findings: (1) Both the rationale score and retrieval score contribute positively to RADIO's performance, with the rationale score demonstrating a stronger positive impact compared to the retrieval score. (2) RADIO outperforms both \textbf{w/o ALL} and \textbf{w/o Retrieval}, while \textbf{w/o Retrieval} surpasses \textbf{w/o ALL}. This indicates that the rationale score and retrieval score are not conflicting but rather complementary. Their integration provides a more robust signal for document ranking, which effectively aids in fine-tuning the reranker. (3) In the STEM category of the MMLU dataset, \textbf{w/o ALL} outperforms both \textbf{w/o Retrieval} and RADIO. This could be attributed to the fact that the training dataset (NQ) contains questions with distributions more similar to humanities and social sciences, leading to trends in the Humanities, Social Sciences, and Other categories that differ from STEM category.

\subsection{Hyperparameter Analysis}

\begin{figure}
    \centering
    \includegraphics[width=\linewidth]{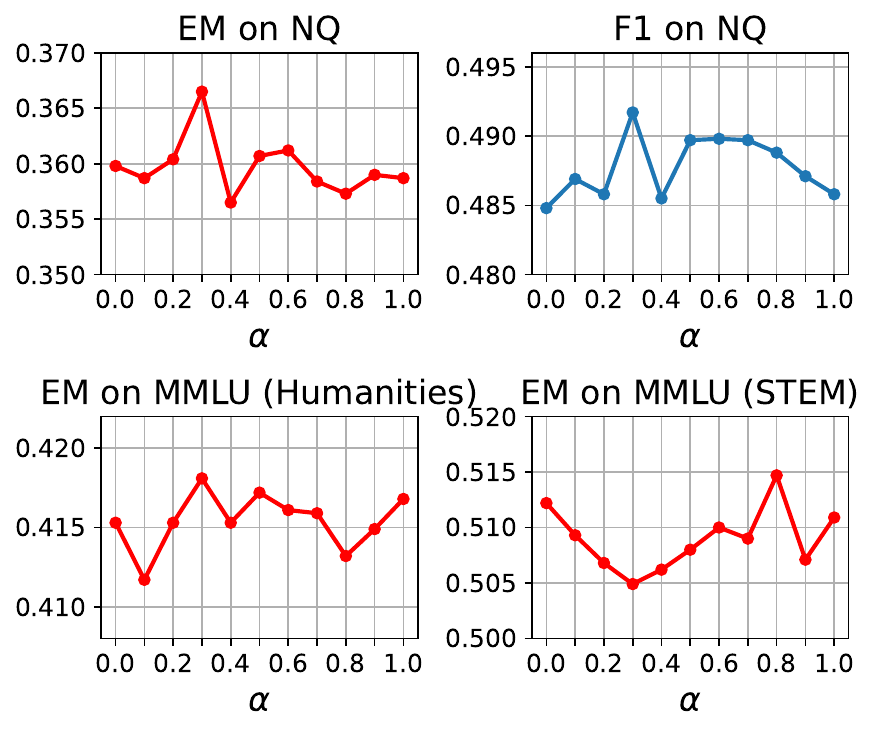}
    \caption{Hyperparameter analysis on NQ and MMLU.}
    \label{fig:hyper-nq}
\end{figure}

Figure~\ref{fig:hyper-nq} visualizes the performance of RADIO across different integration coefficients $\alpha$ on NQ and MMLU. The x-axis represents the integration coefficients $\alpha$ and the y-axis represents the evaluation metrics EM (red) or F1 (blue). For the MMLU dataset, we present results for two representative categories: Humanities and STEM. The trends for other categories align with those observed in Humanities. Complete experimental results are provided in Appendix~\ref{sec:more_hyper} for reference. From the figure, we can draw the following conclusions:
(1) As $\alpha$ increases, the metrics on the NQ dataset and most categories of the MMLU dataset exhibit a trend of first rising and then falling, with the optimal range for $\alpha$ being between 0.3 and 0.7. This demonstrates the complementary nature of rationale scores and retrieval scores, which together form an optimal signal for preference alignment.
(2) When $\alpha=0$, the RAG performance is suboptimal because document scoring relies entirely on retrieval scores, focusing solely on query-document relevance while ignoring whether the document supports the generator in answering the query. Conversely, when $\alpha=1$, the performance is still not optimal, as it completely disregards retrieval relevance, leading to a mismatch between the fine-tuning dataset and the training data, which negatively affects model performance.
(3) The trends for STEM differ from those of other MMLU categories, showing an opposite pattern. This is likely due to the significant distributional differences between STEM and Humanities/Social Sciences, resulting in a "seesaw effect" as observed in the figure. It is worth noting that this phenomenon is also reflected in REPLUG~\cite{shi2023replug}, where the improvement in the STEM category is weaker compared to other categories.

\section{Conclusion}

In this paper, we propose a novel and practical preference alignment framework, RADIO, with rationale distillation in retrieval-augmented generation. First, we introduce a rationale extraction method to extract the rationales necessary for answering queries with LLMs. Next, a rationale-based alignment is proposed to rerank documents based on extracted rationales and fine-tune rerankers. 
Extensive experiments on three tasks across four datasets are conducted to validate the effectiveness of our proposed method against state-of-the-art baselines and demonstrate its strong transferability. We also release our code online to facilitate the following research and ease of reproduction.

\clearpage
\section{Limitations}
First, compared to other methods, our approach RADIO requires additional time in the rationale extraction stage to generate rationales. Since different samples are independent of one another, we can reduce generation time by employing parallel processing to mitigate this issue. Secondly, the MMLU experimental results reveal that the composition of fine-tuning datasets can affect RADIO's effectiveness. This issue can be addressed by designing task-specific fine-tuning datasets for different downstream tasks or large-scale general fine-tuning datasets.
\section*{Acknowledgments}
This research was partially supported by Research Impact Fund (No.R1015-23), Collaborative Research Fund (No.C1043-24GF), and Huawei (Huawei Innovation Research Program, Huawei Fellowship).

\bibliography{reference}

\begin{thebibliography}{46}
\providecommand{\natexlab}[1]{#1}

\bibitem[{Chen et~al.(2024)Chen, Xiao, Zhang, Luo, Lian, and Liu}]{bge-m3}
Jianlv Chen, Shitao Xiao, Peitian Zhang, Kun Luo, Defu Lian, and Zheng Liu. 2024.
\newblock \href {https://arxiv.org/abs/2402.03216} {Bge m3-embedding: Multi-lingual, multi-functionality, multi-granularity text embeddings through self-knowledge distillation}.
\newblock \emph{Preprint}, arXiv:2402.03216.

\bibitem[{Dong et~al.(2024)Dong, Zhu, Zhang, Wang, Dou, and Wen}]{dong2024understand}
Guanting Dong, Yutao Zhu, Chenghao Zhang, Zechen Wang, Zhicheng Dou, and Ji-Rong Wen. 2024.
\newblock Understand what llm needs: Dual preference alignment for retrieval-augmented generation.
\newblock \emph{arXiv preprint arXiv:2406.18676}.

\bibitem[{Fan et~al.(2024)Fan, Ding, Ning, Wang, Li, Yin, Chua, and Li}]{fan2024survey}
Wenqi Fan, Yujuan Ding, Liangbo Ning, Shijie Wang, Hengyun Li, Dawei Yin, Tat-Seng Chua, and Qing Li. 2024.
\newblock A survey on rag meeting llms: Towards retrieval-augmented large language models.
\newblock In \emph{Proceedings of the 30th ACM SIGKDD Conference on Knowledge Discovery and Data Mining}, pages 6491--6501.

\bibitem[{Fu et~al.(2023)Fu, Li, Wu, Wang, Dong, Zhao, Zhao, Guo, and Tang}]{fu2023unified}
Zichuan Fu, Xiangyang Li, Chuhan Wu, Yichao Wang, Kuicai Dong, Xiangyu Zhao, Mengchen Zhao, Huifeng Guo, and Ruiming Tang. 2023.
\newblock A unified framework for multi-domain ctr prediction via large language models.
\newblock \emph{ACM Transactions on Information Systems}.

\bibitem[{Gao et~al.(2023)Gao, Xiong, Gao, Jia, Pan, Bi, Dai, Sun, and Wang}]{gao2023retrieval}
Yunfan Gao, Yun Xiong, Xinyu Gao, Kangxiang Jia, Jinliu Pan, Yuxi Bi, Yi~Dai, Jiawei Sun, and Haofen Wang. 2023.
\newblock Retrieval-augmented generation for large language models: A survey.
\newblock \emph{arXiv preprint arXiv:2312.10997}.

\bibitem[{Hendrycks et~al.(2020)Hendrycks, Burns, Basart, Zou, Mazeika, Song, and Steinhardt}]{hendrycks2020measuring}
Dan Hendrycks, Collin Burns, Steven Basart, Andy Zou, Mantas Mazeika, Dawn Song, and Jacob Steinhardt. 2020.
\newblock Measuring massive multitask language understanding.
\newblock \emph{arXiv preprint arXiv:2009.03300}.

\bibitem[{Izacard et~al.(2023)Izacard, Lewis, Lomeli, Hosseini, Petroni, Schick, Dwivedi-Yu, Joulin, Riedel, and Grave}]{izacard2023atlas}
Gautier Izacard, Patrick Lewis, Maria Lomeli, Lucas Hosseini, Fabio Petroni, Timo Schick, Jane Dwivedi-Yu, Armand Joulin, Sebastian Riedel, and Edouard Grave. 2023.
\newblock Atlas: Few-shot learning with retrieval augmented language models.
\newblock \emph{Journal of Machine Learning Research}, 24(251):1--43.

\bibitem[{Jia et~al.(2025)Jia, Cai, Zhao, Pan, Xin, Huang, Zhang, Zhao, Yin, and Yang}]{jia2025agentir}
Pengyue Jia, Qingpeng Cai, Xiangyu Zhao, Ling Pan, Xin Xin, Jin Huang, Weinan Zhang, Li~Zhao, Dawei Yin, and Grace~Hui Yang. 2025.
\newblock Agentir: 2nd workshop on agent-based information retrieval.
\newblock ACM SIGIR.

\bibitem[{Jia et~al.(2024{\natexlab{a}})Jia, Liu, Li, Zhao, Wang, Du, Han, Wei, Wang, and Yin}]{jia2024g3}
Pengyue Jia, Yiding Liu, Xiaopeng Li, Xiangyu Zhao, Yuhao Wang, Yantong Du, Xiao Han, Xuetao Wei, Shuaiqiang Wang, and Dawei Yin. 2024{\natexlab{a}}.
\newblock G3: an effective and adaptive framework for worldwide geolocalization using large multi-modality models.
\newblock \emph{Advances in Neural Information Processing Systems}, 37:53198--53221.

\bibitem[{Jia et~al.(2024{\natexlab{b}})Jia, Liu, Zhao, Li, Hao, Wang, and Yin}]{jia2024mill}
Pengyue Jia, Yiding Liu, Xiangyu Zhao, Xiaopeng Li, Changying Hao, Shuaiqiang Wang, and Dawei Yin. 2024{\natexlab{b}}.
\newblock Mill: Mutual verification with large language models for zero-shot query expansion.
\newblock In \emph{Proceedings of the 2024 Conference of the North American Chapter of the Association for Computational Linguistics: Human Language Technologies (Volume 1: Long Papers)}, pages 2498--2518.

\bibitem[{Jiang et~al.(2024)Jiang, Chen, Bai, He, Li, Yang, Zhao, Nie, and Zhang}]{jiang2024survey}
Ruili Jiang, Kehai Chen, Xuefeng Bai, Zhixuan He, Juntao Li, Muyun Yang, Tiejun Zhao, Liqiang Nie, and Min Zhang. 2024.
\newblock A survey on human preference learning for large language models.
\newblock \emph{arXiv preprint arXiv:2406.11191}.

\bibitem[{Jin et~al.(2024)Jin, Zhu, Yang, Zhang, and Dou}]{jin2024flashrag}
Jiajie Jin, Yutao Zhu, Xinyu Yang, Chenghao Zhang, and Zhicheng Dou. 2024.
\newblock Flashrag: A modular toolkit for efficient retrieval-augmented generation research.
\newblock \emph{arXiv preprint arXiv:2405.13576}.

\bibitem[{Joshi et~al.(2017)Joshi, Choi, Weld, and Zettlemoyer}]{joshi2017triviaqa}
Mandar Joshi, Eunsol Choi, Daniel~S Weld, and Luke Zettlemoyer. 2017.
\newblock Triviaqa: A large scale distantly supervised challenge dataset for reading comprehension.
\newblock In \emph{Proceedings of the 55th Annual Meeting of the Association for Computational Linguistics (Volume 1: Long Papers)}, pages 1601--1611.

\bibitem[{Kaplan et~al.(2020)Kaplan, McCandlish, Henighan, Brown, Chess, Child, Gray, Radford, Wu, and Amodei}]{kaplan2020scaling}
Jared Kaplan, Sam McCandlish, Tom Henighan, Tom~B Brown, Benjamin Chess, Rewon Child, Scott Gray, Alec Radford, Jeffrey Wu, and Dario Amodei. 2020.
\newblock Scaling laws for neural language models.
\newblock \emph{arXiv preprint arXiv:2001.08361}.

\bibitem[{Ke et~al.(2024)Ke, Kong, Li, Zhang, Mei, and Bendersky}]{ke2024bridging}
Zixuan Ke, Weize Kong, Cheng Li, Mingyang Zhang, Qiaozhu Mei, and Michael Bendersky. 2024.
\newblock Bridging the preference gap between retrievers and llms.
\newblock \emph{arXiv preprint arXiv:2401.06954}.

\bibitem[{Kingma(2014)}]{kingma2014adam}
Diederik~P Kingma. 2014.
\newblock Adam: A method for stochastic optimization.
\newblock \emph{arXiv preprint arXiv:1412.6980}.

\bibitem[{Kwiatkowski et~al.(2019)Kwiatkowski, Palomaki, Redfield, Collins, Parikh, Alberti, Epstein, Polosukhin, Devlin, Lee et~al.}]{kwiatkowski2019natural}
Tom Kwiatkowski, Jennimaria Palomaki, Olivia Redfield, Michael Collins, Ankur Parikh, Chris Alberti, Danielle Epstein, Illia Polosukhin, Jacob Devlin, Kenton Lee, et~al. 2019.
\newblock Natural questions: a benchmark for question answering research.
\newblock \emph{Transactions of the Association for Computational Linguistics}, 7:453--466.

\bibitem[{Li et~al.(2023{\natexlab{a}})Li, Su, Jia, Zhao, Cheng, Wang, and Yin}]{li2023agent4ranking}
Xiaopeng Li, Lixin Su, Pengyue Jia, Xiangyu Zhao, Suqi Cheng, Junfeng Wang, and Dawei Yin. 2023{\natexlab{a}}.
\newblock Agent4ranking: Semantic robust ranking via personalized query rewriting using multi-agent llm.
\newblock \emph{arXiv preprint arXiv:2312.15450}.

\bibitem[{Li et~al.(2023{\natexlab{b}})Li, Zhang, Zhang, Long, Xie, and Zhang}]{li2023towards}
Zehan Li, Xin Zhang, Yanzhao Zhang, Dingkun Long, Pengjun Xie, and Meishan Zhang. 2023{\natexlab{b}}.
\newblock Towards general text embeddings with multi-stage contrastive learning.
\newblock \emph{arXiv preprint arXiv:2308.03281}.

\bibitem[{Liu et~al.(2024)Liu, Wu, Zhao, Zhu, Zhang, Tian, and Zheng}]{liu2024large}
Qidong Liu, Xian Wu, Xiangyu Zhao, Yuanshao Zhu, Zijian Zhang, Feng Tian, and Yefeng Zheng. 2024.
\newblock Large language model distilling medication recommendation model.
\newblock \emph{arXiv preprint arXiv:2402.02803}.

\bibitem[{Ma et~al.(2023)Ma, Gong, He, Zhao, and Duan}]{ma2023query}
Xinbei Ma, Yeyun Gong, Pengcheng He, Hai Zhao, and Nan Duan. 2023.
\newblock Query rewriting for retrieval-augmented large language models.
\newblock \emph{arXiv preprint arXiv:2305.14283}.

\bibitem[{Moreira et~al.(2024{\natexlab{a}})Moreira, Ak, Schifferer, Xu, Osmulski, and Oldridge}]{moreira2024enhancing}
Gabriel de Souza~P Moreira, Ronay Ak, Benedikt Schifferer, Mengyao Xu, Radek Osmulski, and Even Oldridge. 2024{\natexlab{a}}.
\newblock Enhancing q\&a text retrieval with ranking models: Benchmarking, fine-tuning and deploying rerankers for rag.
\newblock \emph{arXiv preprint arXiv:2409.07691}.

\bibitem[{Moreira et~al.(2024{\natexlab{b}})Moreira, Osmulski, Xu, Ak, Schifferer, and Oldridge}]{moreira2024nv}
Gabriel de Souza~P Moreira, Radek Osmulski, Mengyao Xu, Ronay Ak, Benedikt Schifferer, and Even Oldridge. 2024{\natexlab{b}}.
\newblock Nv-retriever: Improving text embedding models with effective hard-negative mining.
\newblock \emph{arXiv preprint arXiv:2407.15831}.

\bibitem[{Ni et~al.(2025)Ni, Zhang, Miao, Zhao, Wu, Wang, and Yin}]{ni2025zeroed}
Wei Ni, Kaihang Zhang, Xiaoye Miao, Xiangyu Zhao, Yangyang Wu, Yaoshu Wang, and Jianwei Yin. 2025.
\newblock Zeroed: Hybrid zero-shot error detection through large language model reasoning.
\newblock \emph{arXiv preprint arXiv:2504.05345}.

\bibitem[{Ouyang et~al.(2022)Ouyang, Wu, Jiang, Almeida, Wainwright, Mishkin, Zhang, Agarwal, Slama, Ray et~al.}]{ouyang2022training}
Long Ouyang, Jeffrey Wu, Xu~Jiang, Diogo Almeida, Carroll Wainwright, Pamela Mishkin, Chong Zhang, Sandhini Agarwal, Katarina Slama, Alex Ray, et~al. 2022.
\newblock Training language models to follow instructions with human feedback.
\newblock \emph{Advances in neural information processing systems}, 35:27730--27744.

\bibitem[{Rafailov et~al.(2024)Rafailov, Sharma, Mitchell, Manning, Ermon, and Finn}]{rafailov2024direct}
Rafael Rafailov, Archit Sharma, Eric Mitchell, Christopher~D Manning, Stefano Ermon, and Chelsea Finn. 2024.
\newblock Direct preference optimization: Your language model is secretly a reward model.
\newblock \emph{Advances in Neural Information Processing Systems}, 36.

\bibitem[{Shi et~al.(2023)Shi, Min, Yasunaga, Seo, James, Lewis, Zettlemoyer, and Yih}]{shi2023replug}
Weijia Shi, Sewon Min, Michihiro Yasunaga, Minjoon Seo, Rich James, Mike Lewis, Luke Zettlemoyer, and Wen-tau Yih. 2023.
\newblock Replug: Retrieval-augmented black-box language models.
\newblock \emph{arXiv preprint arXiv:2301.12652}.

\bibitem[{Singh et~al.(2021)Singh, Reddy, Hamilton, Dyer, and Yogatama}]{singh2021end}
Devendra Singh, Siva Reddy, Will Hamilton, Chris Dyer, and Dani Yogatama. 2021.
\newblock End-to-end training of multi-document reader and retriever for open-domain question answering.
\newblock \emph{Advances in Neural Information Processing Systems}, 34:25968--25981.

\bibitem[{Touvron et~al.(2023)Touvron, Lavril, Izacard, Martinet, Lachaux, Lacroix, Rozi{\`e}re, Goyal, Hambro, Azhar et~al.}]{touvron2023llama}
Hugo Touvron, Thibaut Lavril, Gautier Izacard, Xavier Martinet, Marie-Anne Lachaux, Timoth{\'e}e Lacroix, Baptiste Rozi{\`e}re, Naman Goyal, Eric Hambro, Faisal Azhar, et~al. 2023.
\newblock Llama: Open and efficient foundation language models.
\newblock \emph{arXiv preprint arXiv:2302.13971}.

\bibitem[{Trivedi et~al.(2022)Trivedi, Balasubramanian, Khot, and Sabharwal}]{trivedi2022musique}
Harsh Trivedi, Niranjan Balasubramanian, Tushar Khot, and Ashish Sabharwal. 2022.
\newblock Musique: Multihop questions via single-hop question composition.
\newblock \emph{Transactions of the Association for Computational Linguistics}, 10:539--554.

\bibitem[{Wang et~al.(2022)Wang, Yang, Huang, Jiao, Yang, Jiang, Majumder, and Wei}]{wang2022text}
Liang Wang, Nan Yang, Xiaolong Huang, Binxing Jiao, Linjun Yang, Daxin Jiang, Rangan Majumder, and Furu Wei. 2022.
\newblock Text embeddings by weakly-supervised contrastive pre-training.
\newblock \emph{arXiv preprint arXiv:2212.03533}.

\bibitem[{Wang et~al.(2023{\natexlab{a}})Wang, Yang, and Wei}]{wang2023query2doc}
Liang Wang, Nan Yang, and Furu Wei. 2023{\natexlab{a}}.
\newblock Query2doc: Query expansion with large language models.
\newblock In \emph{Proceedings of the 2023 Conference on Empirical Methods in Natural Language Processing}, pages 9414--9423.

\bibitem[{Wang et~al.(2023{\natexlab{b}})Wang, Zhao, Liu, Chen, Zhuang, Gu, Guo, and Zhao}]{wang2023large}
Maolin Wang, Yao Zhao, Jiajia Liu, Jingdong Chen, Chenyi Zhuang, Jinjie Gu, Ruocheng Guo, and Xiangyu Zhao. 2023{\natexlab{b}}.
\newblock Large multimodal model compression via efficient pruning and distillation at antgroup.
\newblock \emph{arXiv preprint arXiv:2312.05795}.

\bibitem[{Wei et~al.(2022)Wei, Wang, Schuurmans, Bosma, Xia, Chi, Le, Zhou et~al.}]{wei2022chain}
Jason Wei, Xuezhi Wang, Dale Schuurmans, Maarten Bosma, Fei Xia, Ed~Chi, Quoc~V Le, Denny Zhou, et~al. 2022.
\newblock Chain-of-thought prompting elicits reasoning in large language models.
\newblock \emph{Advances in neural information processing systems}, 35:24824--24837.

\bibitem[{Xiao et~al.(2023)Xiao, Liu, Zhang, and Muennighoff}]{bge_embedding}
Shitao Xiao, Zheng Liu, Peitian Zhang, and Niklas Muennighoff. 2023.
\newblock \href {https://arxiv.org/abs/2309.07597} {C-pack: Packaged resources to advance general chinese embedding}.
\newblock \emph{Preprint}, arXiv:2309.07597.

\bibitem[{Xu et~al.(2025{\natexlab{a}})Xu, Jia, Li, Zhang, Wang, Liu, Zhao, Wang, Guo, Tang et~al.}]{xu2025align}
Derong Xu, Pengyue Jia, Xiaopeng Li, Yingyi Zhang, Maolin Wang, Qidong Liu, Xiangyu Zhao, Yichao Wang, Huifeng Guo, Ruiming Tang, et~al. 2025{\natexlab{a}}.
\newblock Align-grag: Reasoning-guided dual alignment for graph retrieval-augmented generation.
\newblock \emph{arXiv preprint arXiv:2505.16237}.

\bibitem[{Xu et~al.(2025{\natexlab{b}})Xu, Li, Zhang, Lin, Zhu, Zheng, Wu, Zhao, Xu, and Chen}]{xu2025harnessing}
Derong Xu, Xinhang Li, Ziheng Zhang, Zhenxi Lin, Zhihong Zhu, Zhi Zheng, Xian Wu, Xiangyu Zhao, Tong Xu, and Enhong Chen. 2025{\natexlab{b}}.
\newblock Harnessing large language models for knowledge graph question answering via adaptive multi-aspect retrieval-augmentation.
\newblock In \emph{Proceedings of the AAAI Conference on Artificial Intelligence}, volume~39, pages 25570--25578.

\bibitem[{Xu et~al.(2024)Xu, Zhang, Lin, Wu, Zhu, Xu, Zhao, Zheng, and Chen}]{xu2024multi}
Derong Xu, Ziheng Zhang, Zhenxi Lin, Xian Wu, Zhihong Zhu, Tong Xu, Xiangyu Zhao, Yefeng Zheng, and Enhong Chen. 2024.
\newblock Multi-perspective improvement of knowledge graph completion with large language models.
\newblock \emph{arXiv preprint arXiv:2403.01972}.

\bibitem[{Yang et~al.(2024)Yang, Yang, Hui, Zheng, Yu, Zhou, Li, Li, Liu, Huang, Dong, Wei, Lin, Tang, Wang, Yang, Tu, Zhang, Ma, Xu, Zhou, Bai, He, Lin, Dang, Lu, Chen, Yang, Li, Xue, Ni, Zhang, Wang, Peng, Men, Gao, Lin, Wang, Bai, Tan, Zhu, Li, Liu, Ge, Deng, Zhou, Ren, Zhang, Wei, Ren, Fan, Yao, Zhang, Wan, Chu, Liu, Cui, Zhang, and Fan}]{qwen2}
An~Yang, Baosong Yang, Binyuan Hui, Bo~Zheng, Bowen Yu, Chang Zhou, Chengpeng Li, Chengyuan Li, Dayiheng Liu, Fei Huang, Guanting Dong, Haoran Wei, Huan Lin, Jialong Tang, Jialin Wang, Jian Yang, Jianhong Tu, Jianwei Zhang, Jianxin Ma, Jin Xu, Jingren Zhou, Jinze Bai, Jinzheng He, Junyang Lin, Kai Dang, Keming Lu, Keqin Chen, Kexin Yang, Mei Li, Mingfeng Xue, Na~Ni, Pei Zhang, Peng Wang, Ru~Peng, Rui Men, Ruize Gao, Runji Lin, Shijie Wang, Shuai Bai, Sinan Tan, Tianhang Zhu, Tianhao Li, Tianyu Liu, Wenbin Ge, Xiaodong Deng, Xiaohuan Zhou, Xingzhang Ren, Xinyu Zhang, Xipin Wei, Xuancheng Ren, Yang Fan, Yang Yao, Yichang Zhang, Yu~Wan, Yunfei Chu, Yuqiong Liu, Zeyu Cui, Zhenru Zhang, and Zhihao Fan. 2024.
\newblock Qwen2 technical report.
\newblock \emph{arXiv preprint arXiv:2407.10671}.

\bibitem[{Yao et~al.(2022)Yao, Zhao, Yu, Du, Shafran, Narasimhan, and Cao}]{yao2022react}
Shunyu Yao, Jeffrey Zhao, Dian Yu, Nan Du, Izhak Shafran, Karthik Narasimhan, and Yuan Cao. 2022.
\newblock React: Synergizing reasoning and acting in language models.
\newblock \emph{arXiv preprint arXiv:2210.03629}.

\bibitem[{Yu et~al.(2023)Yu, Xiong, Yu, and Liu}]{yu2023augmentation}
Zichun Yu, Chenyan Xiong, Shi Yu, and Zhiyuan Liu. 2023.
\newblock Augmentation-adapted retriever improves generalization of language models as generic plug-in.
\newblock \emph{arXiv preprint arXiv:2305.17331}.

\bibitem[{Zhang et~al.(2024)Zhang, Yu, Wang, and Zhang}]{zhang2024arl2}
Lingxi Zhang, Yue Yu, Kuan Wang, and Chao Zhang. 2024.
\newblock Arl2: Aligning retrievers for black-box large language models via self-guided adaptive relevance labeling.
\newblock \emph{arXiv preprint arXiv:2402.13542}.

\bibitem[{Zhang et~al.(2025)Zhang, Jia, Li, Xu, Wang, Wang, Du, Guo, Liu, Tang et~al.}]{zhang2025lsrp}
Yingyi Zhang, Pengyue Jia, Xianneng Li, Derong Xu, Maolin Wang, Yichao Wang, Zhaocheng Du, Huifeng Guo, Yong Liu, Ruiming Tang, et~al. 2025.
\newblock Lsrp: A leader-subordinate retrieval framework for privacy-preserving cloud-device collaboration.
\newblock \emph{arXiv preprint arXiv:2505.05031}.

\bibitem[{Zhao et~al.(2023)Zhao, Zhou, Li, Tang, Wang, Hou, Min, Zhang, Zhang, Dong et~al.}]{zhao2023survey}
Wayne~Xin Zhao, Kun Zhou, Junyi Li, Tianyi Tang, Xiaolei Wang, Yupeng Hou, Yingqian Min, Beichen Zhang, Junjie Zhang, Zican Dong, et~al. 2023.
\newblock A survey of large language models.
\newblock \emph{arXiv preprint arXiv:2303.18223}.

\bibitem[{Zhao et~al.(2018{\natexlab{a}})Zhao, Xia, Zhang, Ding, Yin, and Tang}]{zhao2018deep}
Xiangyu Zhao, Long Xia, Liang Zhang, Zhuoye Ding, Dawei Yin, and Jiliang Tang. 2018{\natexlab{a}}.
\newblock Deep reinforcement learning for page-wise recommendations.
\newblock In \emph{Proceedings of the 12th ACM conference on recommender systems}, pages 95--103.

\bibitem[{Zhao et~al.(2018{\natexlab{b}})Zhao, Zhang, Ding, Xia, Tang, and Yin}]{zhao2018recommendations}
Xiangyu Zhao, Liang Zhang, Zhuoye Ding, Long Xia, Jiliang Tang, and Dawei Yin. 2018{\natexlab{b}}.
\newblock Recommendations with negative feedback via pairwise deep reinforcement learning.
\newblock In \emph{Proceedings of the 24th ACM SIGKDD international conference on knowledge discovery \& data mining}, pages 1040--1048.

\end{thebibliography}

\clearpage
\appendix
\section{Appendix}

\subsection{Dataset Desctiptions} \label{sec:appendix_dataset}

The detailed descriptions of baselines are given as follows:
\begin{itemize}
    \item \textbf{Natural Questions (NQ):} NQ contains real user questions compiled from Google search, with corresponding answers identified from Wikipedia by human annotators.
    \item \textbf{TriviaQA:} TriviaQA dataset comprises trivia questions paired with answer annotations and supporting evidence documents, such as web pages and Wikipedia articles. It is designed to assess a model's ability to retrieve and comprehend textual evidence for open-domain question answering.
    \item \textbf{Massive Multitask Language Understanding (MMLU):} MMLU is a comprehensive evaluation dataset comprising 57 categories of questions, which are grouped into four broad domains: Humanities, Social Sciences, STEM, and Other. In this paper, we report evaluation metrics based on these categories.
    \item \textbf{Musique:} MuSiQue is a multi-hop question answering dataset created using a bottom-up approach to ensure connected reasoning, with 25,000 2–4 hop questions.
\end{itemize}

\subsection{Baselines} \label{sec:appendix_baseline}

Following is the introduction of baselines:
\begin{itemize}
    \item \textbf{Base~\cite{bge_embedding}:} The reranker model is used off-the-shelf without any fine-tuning.
    \item \textbf{Atlas~\cite{izacard2023atlas}:} A pretrained retrieval-augmented language model designed for knowledge intensive task. We choose the EMDR$^2$~\cite{singh2021end} as the reward to rerank documents and fine-tune rerankers.
    \item \textbf{REPLUG~\cite{shi2023replug}:} REPLUG seeks to fine-tune the retriever to enhance RAG pipelines that include black-box LLMs. It achieves this by using the query and document as contextual inputs and leveraging the probability of the LLM generating the correct answer as the importance score.
    This idea is also reflected in the PDist method in Atlas~\cite{izacard2023atlas}.
    \item \textbf{Trainable rewrite-retrieve-read (RRR)~\cite{ma2023query}:} RRR optimizes the query rewriter using the evaluation metrics of the final RAG output as a reward, which is used to fine-tune rerankers in our pipeline, enhancing the overall effectiveness of RAG.
    \item \textbf{ARL2~\cite{zhang2024arl2}:} ARL2 introduces a method to use LLMs as supervisor to generate self-guided relevance labels (e.g., ``Not relevant'', ``relevant'') for fine-tuning retriever.
    \item \textbf{DPA-RAG~\cite{dong2024understand}:} DPA-RAG proposes a knowledge preference pipeline to dual-align rerankers and generators in RAG. It combines document importance from the LLM's perspective with the importance determined during the retrieval stage.
\end{itemize}

\subsection{Detailed Introduction to Backbone Rerankers} \label{sec:appendix_reranker}

To validate the generality and adaptability of RADIO, we select three different rerankers as the backbone models for our experiments: (1) \textbf{gte-base}~\cite{li2023towards}: a reranker model proposed by Alibaba DAMO Academy, with 109M parameters and 768 embedding dimensions. (2) \textbf{gte-large}~\cite{li2023towards}: the larger version of gte-base, with 335M parameters and 1024 embedding dimensions. (3) \textbf{bge-reranker-base}~\cite{bge_embedding}: a powerful cross-encoder architecture reranker proposed by Beijing Academy of Artificial Intelligence, with 278M parameters.

\subsection{Prompts} \label{sec:appendix_prompts}

\begin{table}
\centering
\caption{Prompts for QA datasets.}
\label{tab:appendix_prompt_open_domain}
\resizebox{0.95\linewidth}{!}{
\begin{tabular}{l} 
\toprule
\begin{tabular}[c]{@{}l@{}}\textbf{System Prompt}: Answer the question based on \\the given document. Only give me the answer \\and do not output any other words. The\\following are given documents.\\\{reference\}\\\textbf{User Prompt}:~\\Question: \{question\}\\Answer:\end{tabular}  \\
\bottomrule
\end{tabular}}
\end{table}

\begin{table}
\centering
\caption{Prompts for Multi-choice datasets.}
\label{tab:appendix_prompt_multi_choice}
\resizebox{0.95\linewidth}{!}{
\begin{tabular}{l} 
\toprule
\begin{tabular}[c]{@{}l@{}}\textbf{System Prompt}: Answer the question based on \\the given document. Only give me the option~\\(A/B/C/D) and do not output any other words. \\The following are given documents.\\\{reference\}\\\textbf{User Prompt}:~\\Question: \{question\}\\Answer:\end{tabular}  \\
\bottomrule
\end{tabular}}
\end{table}

In this section, we detail the prompts we used in the experiments.
For Open-domain QA datasets (NQ and TriviaQA) and Multi-hop QA dataset (Musique), the prompts are as shown in Table~\ref{tab:appendix_prompt_open_domain}. For Multi-choice dataset MMLU, the prompts are shown in Table~\ref{tab:appendix_prompt_multi_choice}.

\subsection{Sample Efficiency Analysis of Fine-tuning} \label{sec:appendix_sample}

To understand the impact of fine-tuning samples on RAG performance, we conduct additional experiments to evaluate RADIO's performance under a few-shot learning setup using varying amounts of fine-tuning samples (0, 4,000, 8,000, 12,000, 16,000, and 20,000 instances) on the NQ dataset with \textit{bge-reranker-base}. The results are shown in Table~\ref{table:appendix_sample}.

\begin{table}
\centering
\caption{Experimental results on NQ with different numbers of fine-tuning samples.}
\label{table:appendix_sample}
\resizebox{0.7\linewidth}{!}{
\begin{tabular}{ccc} 
\toprule
Sample Num. & EM     & F1      \\ 
\midrule
0           & 0.3371 & 0.4603  \\
4,000       & 0.3526 & 0.4843  \\
8,000       & 0.3634 & 0.4896  \\
12,000      & 0.3582 & 0.4842  \\
16,000      & 0.3604 & 0.4889  \\
20,000      & 0.3665 & 0.4917  \\
\bottomrule
\end{tabular}}
\end{table}

We observe that even with a relatively small number of fine-tuning samples (e.g., 4,000 or 8,000), RADIO yields substantial improvements over the zero-shot baseline. Notably, with 8,000 samples the EM score increases by 7.8\% and the F1 score by 6.4\%, clearly demonstrating the model's capacity to benefit from limited task-specific data. Moreover, as the number of fine-tuning instances increases, the overall performance tends to improve, suggesting that additional annotated data can further enhance the retrieval and generation capabilities of the system. Although there are minor fluctuations—such as a slight dip in performance at 12,000 and 16,000 samples—these variations may stem from differences in data distribution or optimization dynamics.

\subsection{More Analysis on Training Hyperparameters} \label{sec:appendix_m_n}

To understand the specific impact of two training hyperparameters (i.e., the shift $n$ in \textit{Top-k shifted by N} and the number of negative samples $m$) on performance, we conduct the following experiments on NQ based on \textit{bge-reranker-base}:

\begin{itemize}
    \item \textbf{Varying $m$}: We fix $n=3$ and vary $m$ over the values 2,4,6, and 8.
    \item \textbf{Varying $n$}: We fix $m=6$ and vary $n$ over the values 1,2,3,4, and 5.
\end{itemize}

The detailed experimental results are presented in Table~\ref{table:appendix_varying_m} and Table~\ref{table:appendix_varying_n}, respectively. From the results, we can find the following points: 
(1) \textbf{Impact of Negative Samples $m$}: Variations in the number of negative samples $m$ have a relatively modest effect on RADIO's performance. The results suggest that the model is robust to changes in $m$, as only marginal performance differences are observed when $m$ is varied from 2 to 8.
(2) \textbf{Impact of the shift parameter $n$}: In contrast, the shift parameter $n$ in the \textit{Top-k shifted by N} mechanism plays a more critical role. Our experiments show that smaller values of $n$ yield better performance, while larger $n$ values lead to a notable drop in both EM and F1 scores. We hypothesize that as $n$ increases, the task of mining hard negatives becomes more challenging, which in turn hampers the fine-tuning process of RADIO.

\begin{table}
\centering
\caption{Hyperparameter analysis on the number of negative samples $m$.}
\label{table:appendix_varying_m}
\resizebox{0.95\linewidth}{!}{
\begin{tabular}{cccc} 
\toprule
Shift ($n$) & Negative Samples~$m$ & EM     & F1      \\ 
\midrule
3           & 2                    & 0.3615 & 0.4886  \\
3           & 4                    & 0.3601 & 0.4888  \\
3           & 6                    & 0.3665 & 0.4917  \\
3           & 8                    & 0.3604 & 0.4895  \\
\bottomrule
\end{tabular}}
\end{table}

\begin{figure}[t]
    \centering
    \includegraphics[width=0.95\linewidth]{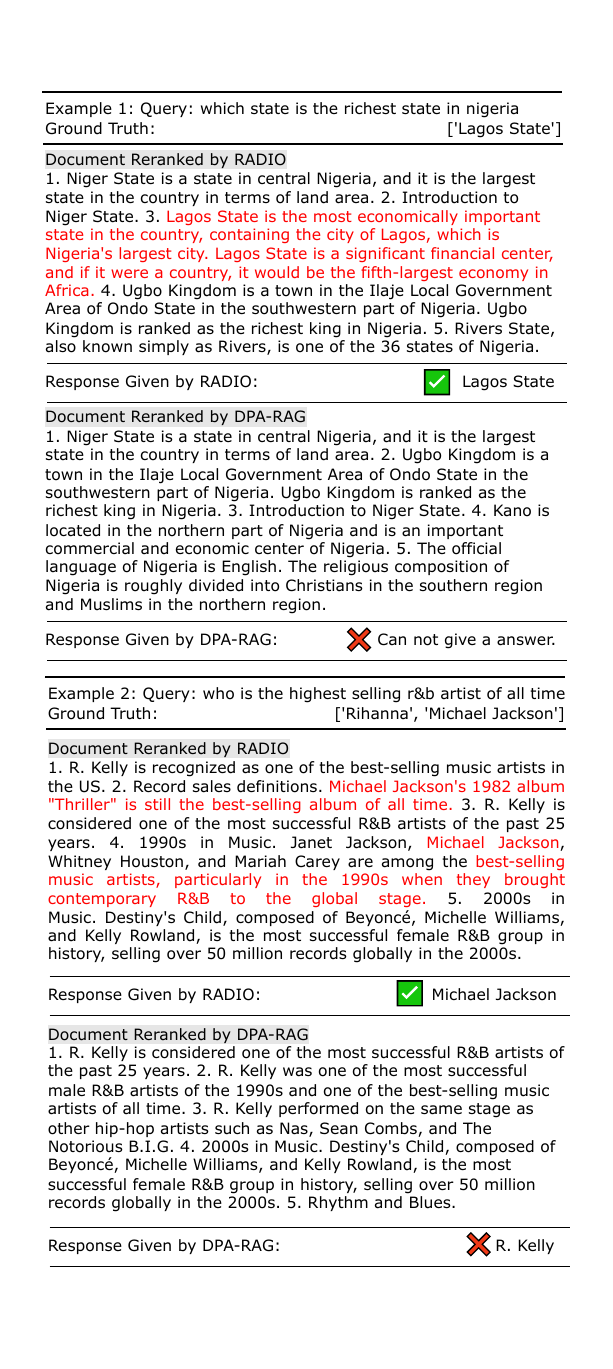}
    \caption{Case study on NQ dataset.}
    \label{fig:case_study}
\end{figure}

\begin{table}
\centering
\caption{Hyperparameter analysis on the shift $n$ in \textit{Top-k shifted by N}}
\label{table:appendix_varying_n}
\resizebox{0.95\linewidth}{!}{
\begin{tabular}{cccc} 
\toprule
Shift ($n$) & Negative Samples~$m$ & EM     & F1      \\ 
\midrule
1           & 6                    & 0.3654 & 0.4918  \\
2           & 6                    & 0.3554 & 0.4826  \\
3           & 6                    & 0.3665 & 0.4917  \\
4           & 6                    & 0.3579 & 0.4868  \\
5           & 6                    & 0.3587 & 0.4887  \\
\bottomrule
\end{tabular}}
\end{table}

\subsection{Case Study} \label{sec:appendix_case}

To intuitively illustrate the effectiveness of RADIO, we select examples from the NQ dataset to compare the documents reranked by RADIO with those reranked by a novel baseline, DPA-RAG, as well as their responses.
In Figure~\ref{fig:case_study} Example 1, the query asks, "\textit{Which state is the richest state in Nigeria?}" RADIO successfully ranks information about Lagos State's economic and financial status, which relates to the correct answer, among the top-3 documents. In contrast, DPA-RAG fails to identify documents relevant to answering the query, and cannot provide a valid response. In Figure~\ref{fig:case_study}  Example 2, the query is, "\textit{Who is the highest-selling R\&B artist of all time?}" RADIO prioritizes documents containing information about the correct answer, Michael Jackson, and effectively highlights key terms such as "R\&B" and "best-selling." However, DPA-RAG misinterprets the query's constraints, retrieving documents that either overlook the R\&B artist specification or fail to consider the time span, resulting in an incorrect response. 
These examples demonstrate that RADIO enhances RAG by providing a more efficient and accurate reranking. It selects contextually appropriate documents, enabling the generator to infer correct answers.

\subsection{More Results on MMLU} \label{sec:appendix_mmlu}

In this section, we give the complete experimental results on MMLU. Specifically, we use the NQ dataset and TriviaQA dataset as source dataset to fine-tune rerankers and evaluate them in MMLU. The results are shown in Table~\ref{table:appendix_mmlu}.

\begin{table*}[ht]
\centering
\caption{Experimental results on MMLU. EM is reported as the metric. The source datasets used to fine-tune rerankers are Open-domain QA datasets NQ and TriviaQA.}
\label{table:appendix_mmlu}
\resizebox{\textwidth}{!}{
\begin{tabular}{c|CCCCC|CCCCC} 
\toprule
\multirow{2}{*}{Method} & \multicolumn{5}{c|}{MMLU (Source Dataset NQ)}                                           & \multicolumn{5}{c}{MMLU (Source Dataset TriviaQA)}                                      \\ 
\cmidrule{2-11}
                        & Humanities      & Social          & STEM            & Other           & ALL             & Humanities     & Social          & STEM            & Other           & ALL              \\ 
\midrule
Base                    & 0.4089          & 0.6867          & \textbf{0.5147} & 0.6650          & 0.5502          & 0.4089         & 0.6867          & \textbf{0.5147} & 0.6650          & 0.5502           \\
Atlas                   & 0.3985          & 0.6935          & 0.5074          & 0.6563          & 0.5447          & 0.3966         & 0.6822          & 0.4868          & 0.654           & 0.5364           \\
REPLUG                  & 0.4102          & 0.6854          & 0.5065          & 0.6590          & 0.5473          & 0.3977         & 0.6744          & 0.4821          & 0.6466          & 0.5323           \\
RRR                     & 0.4079          & 0.6913          & \uline{0.5116}  & 0.6572          & 0.5484          & 0.4132         & 0.6926          & 0.5005          & 0.6501          & 0.5464           \\
ARL2                    & 0.4147          & \uline{0.7016}  & 0.5106          & 0.6630          & 0.5540          & 0.4012         & \textbf{0.6951} & 0.5011          & 0.6639          & 0.5462           \\
DPA-RAG                 & \uline{0.4157}  & 0.701           & 0.5078          & \uline{0.6652}  & \uline{0.5541}  & \uline{0.4189} & 0.6932          & 0.5062          & \textbf{0.6746} & \uline{0.5552}   \\
RADIO (Ours)            & \textbf{0.4172} & \textbf{0.7013} & 0.5080          & \textbf{0.6717} & \textbf{0.5562} & \textbf{0.4230} & \uline{0.6942}  & \uline{0.5090}  & \uline{0.6678}  & \textbf{0.5559}  \\
\bottomrule
\end{tabular}}
\end{table*}

\subsection{More Results on Hyperparameter Analysis} \label{sec:more_hyper}

\begin{figure*}[t]
    \centering
    \includegraphics[width=\textwidth]{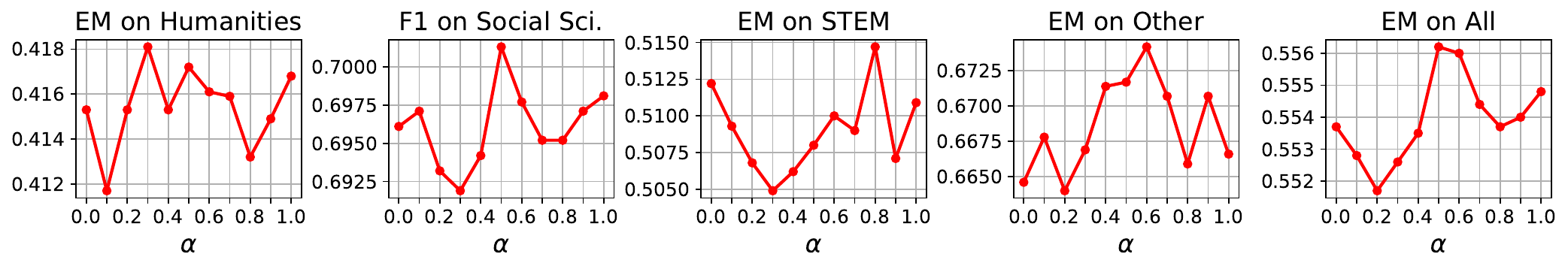}
    \caption{More results of hyperparameter analysis on MMLU.}
    \label{fig:more_hyper}
\end{figure*}

Figure~\ref{fig:more_hyper} illustrates the trend of RADIO's performance across various MMLU categories as the integration coefficient $\alpha$ increases. For Humanities, Social Sciences, and Other, the trends are consistent: as $\alpha$ increases, the performance metrics first improve and then decline. However, the STEM category shows a unique pattern, with metrics initially decreasing as $\alpha$ grows, followed by an improvement. This divergence may be attributed to the fine-tuning dataset (NQ), which shares a closer distribution with Humanities, Social Sciences, and Other categories, while differing significantly from the STEM category.

\end{document}